\documentclass[sigconf, nonacm]{acmart}
\AtBeginDocument{%
  }

\usepackage{multirow} 
\usepackage{xcolor}
\usepackage{graphicx}
\usepackage{booktabs}

\setcopyright{none}
\copyrightyear{2026}
\acmYear{2026}
\acmDOI{XXXXXXX.XXXXXXX}
\acmConference[SIGIR '26]{Proceedings of the SIGIR Conference 2026}{July 20--24, 2026}{Melbourne, Australia}
\acmISBN{}

\begin{document}

\title{DSL: Understanding and Improving Softmax Recommender Systems with Competition-Aware Scaling }

\author{Bucher Sahyouni}
\affiliation{%
  \institution{University of Surrey}
  \city{Guildford}
  \country{United Kingdom}
}
\email{bs00826@surrey.ac.uk}

\author{Matthew Vowels}
\affiliation{%
  \institution{The Sense, CHUV}
  \city{Lausanne}
  \country{Switzerland}
}
\affiliation{%
  \institution{Kivira Health}
  \city{New York}
  \state{NY}
  \country{USA}
}
\email{matthew.vowels@unil.ch}

\author{Liqun Chen}
\affiliation{%
  \institution{University of Surrey}
  \city{Guildford}
  \country{United Kingdom}
}
\email{liqun.chen@surrey.ac.uk}

\author{Simon Hadfield}
\affiliation{%
  \institution{University of Surrey}
  \city{Guildford}
  \country{United Kingdom}
}
\email{s.hadfield@surrey.ac.uk}
\renewcommand{\shortauthors}{Sahyouni et al.}

\begin{abstract}
Softmax Loss (SL) is being increasingly adopted for recommender systems (RS) as it has demonstrated better performance, robustness and fairness. Yet in implicit-feedback, a single global temperature and equal treatment of uniformly sampled negatives can lead to brittle training, because sampled sets may contain varying degrees of relevant or informative competitors. The optimal loss sharpness for a user-item pair with a particular set of negatives, can be suboptimal or destabilising for another with different negatives. 
We introduce \emph{Dual-scale Softmax Loss (DSL)}, which \emph{infers} effective sharpness from the sampled competition itself. DSL adds two complementary branches to the log-sum-exp backbone. Firstly it reweights negatives \emph{within} each training instance using hardness and item--item similarity, secondly it adapts a \emph{per-example} temperature from the competition intensity over a constructed competitor slate. Together, these components preserve the geometry of SL while reshaping the competition distribution across negatives and across examples.

Over several representative benchmarks and backbones, DSL yields substantial gains over strong baselines, with improvements over SL exceeding $10\%$ in several settings and averaging $6.22\%$ across datasets, metrics, and backbones. Under out-of-distribution (OOD) popularity shift, the gains are larger, with an average of $9.31\%$ improvement over SL. We further provide a theoretical, distributionally robust optimisation (DRO) analysis, which demonstrates how DSL reshapes the robust payoff and the KL deviation for ambiguous instances. This helps explain the empirically observed improvements in accuracy and robustness.
\end{abstract}

\ccsdesc[300]{Information systems~Personalization}

\ccsdesc[500]{Information systems~Recommender systems}
\ccsdesc[300]{Information systems~Similarity measures}

\keywords{Recommender systems, loss function, softmax, competitor aware}


\maketitle

\section{Introduction}

\begin{figure}[htbp]
\centering
\includegraphics[width=0.95\columnwidth]{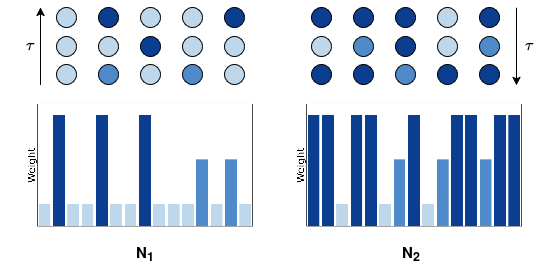}
\includegraphics[width=0.95\columnwidth]{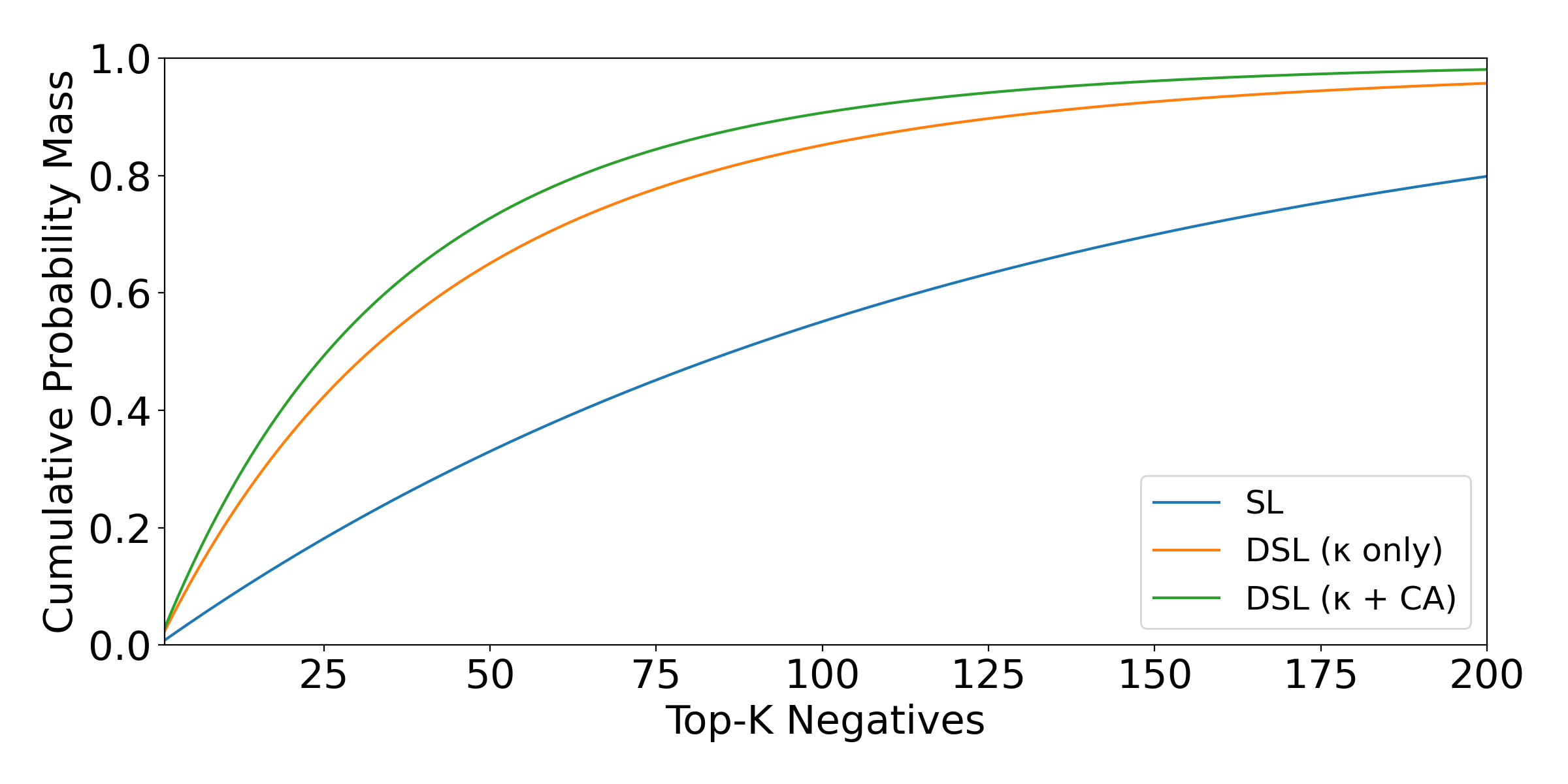}
\caption {Simplified diagram illustrating how DSL reshapes SL competition over two slates. $N_1$ and $N_2$ are two different sets of 15 randomly sampled negatives. $N_1$ has few strong sampled competitors so the per-example $\tau$ (shown with arrows) smoothens the loss. $N_2$ has high competition so the per-example $\tau$ sharpens DSL. The graph shows that within each set, competitive items receive larger relative weightings.}
\label{teaser}
\vspace{-0.4cm}
\end{figure}

Recommender Systems (RS) are at the core of personalised digital platforms \cite{wu2024effectiveness}. Recent advancements have focused on innovations in model architecture. Latent factor models, deep sequential models using transformer self-attention, and graph-based neural network (GNN) recommender systems have been developed to understand and analyse user preferences and relationships more accurately \cite{koren2009matrix,kang2018self_sasrec,He2017WWW_NCF, he2020lightgcn}. On the other hand, the learning objective is often treated as a largely interchangeable component, with many systems defaulting to pointwise or pairwise losses. Recent analyses argue that loss functions in RS can substantially affect robustness, bias, and ranking quality, and have remained comparatively under-explored relative to architectures \cite{wu2024effectiveness, wu2024bsl}. RS objectives are commonly grouped into three categories: \textit{pointwise} losses that score user–item interactions independently, promoting alignment between models predictions and labels (e.g., regression or binary classification), \textit{pairwise} losses that encourage a higher margin between the positive item and sampled negatives (e.g., Bayesian Personalized Ranking (BPR)), and \textit{softmax} losses that normalise over item predictions, creating a multinomial distribution and optimising the probability of positives versus negatives \cite{wu2024effectiveness}. The softmax loss (SL) becomes especially attractive at web scale via sampled softmax (SSM), which approximates the full softmax over a large collection of items by contrasting positives against a sampled negative set. This approach is used in industrial recommendation pipelines and is closely related to contrastive formulations such as InfoNCE \cite{oord2018CPCrepresentation}.

The SL is highly sensitive to the temperature, \(\tau\), which plays a direct role in controlling how sharply the loss concentrates probability over the negatives and therefore the gradient mass assigned to competitors. Too small a \(\tau\) focuses training on fewer negatives and can destabilise training, whereas too high a \(\tau\) could wash out informative competition. Recent analyses discuss the conceptual advantages of SL, such as better hard-negative mining and popularity bias mitigation, but also highlight the dependence of SL performance on how negatives are sampled and weighted \cite{wu2024bsl,wu2024effectiveness}. SL assumes that randomly drawn negatives form a meaningful choice set and treats them equally. Since most sampled items are not plausible substitutes, this diverts gradients away from strong and relevant alternatives. While $\tau$ controls sharpness over the sampled set, it cannot make the pool itself align with true competitors. These observations suggest that a single global \(\tau\) may not be an effective/robust choice in practice, especially when uniformly sampling a set of negatives from a large item collection under implicit feedback data, where non-interactions are confounded by exposure and can contain false negatives \cite{saito2019towards_mnar,marlin2009collaborative_MNAR,schnabel2016recommendations_treatments}.

Our proposed Dual-scale Softmax Loss (DSL) addresses this gap by treating the effective temperature not as a single hyperparameter, but as an inferred quantity which is shaped both by the negatives sampled during training as well as how sharply the loss concentrates gradients on the competitors. To this end, DSL implements two light-weight adaptations into the SL backbone: \emph{(i)} a branch that reweights sampled negatives per example to emphasise meaningful competitors and \emph{(ii)} a competitor-aware (CA) branch that forms a pseudoslate of top competitors and adapts the sharpness per example by deriving a temperature multiplier from how strongly hardness aligns with relevance, such that effective sharpness is shaped by the competition pool. The two branches are implemented with appropriate normalisation to avoid rescaling logits globally i.e., they redistribute weights within-  and per-example. Together, these adaptations attempt to correct the loss's default assumption that all sampled negatives and all examples should be treated equally. Figure~\ref{teaser} illustrates this intuition and how DSL reallocates the learning signal across and within sampled negatives. This weighting behaviour also targets a more fundamental limitation of SL. 
When training recommender systems with implicit feedback data, the sampled softmax loss training regime can be thought of as an approximation to discrete choice models (DCMs), where a user chooses an item from a displayed set and the multinomial logit formulation turns that into choice-from-a-slate probabilities \cite{mcfadden1972conditional}. Unfortunately, the more common implicit feedback datasets lack exposure logs. As such, an absence of interaction could mean either that the user was exposed to the item but did not select it, or that the user was never shown the item. This leads to what is described in the literature as missing-not-at-random (MNAR) exposure bias \cite{saito2019towards_mnar,schnabel2016recommendations_treatments,marlin2009collaborative_MNAR}. The common solution to this is to sample negatives uniformly (or under some guided sampling). However, this approach often picks negatives that were never true competitors and were not co-exposed with the interacted item. This weakens the interpretation of the loss as a true slate conditioned choice. This mismatch motivates DSL. Without impression logs, DSL reshapes the sampled competition signal by emphasising hard and relevant alternatives and adapting the temperature across examples.

We summarise our contributions as follows:
\begin{itemize}
    \item We introduce per-negative reweighting, driven by hardness and relevance proxies, with normalisation per example.
    \item We incorporate per-example temperature adaptation over pseudoslates of top sampled competitors with normalisation across examples.
    \item We conduct experiments on several model backbones and real-world datasets demonstrating the broad applicability of DSL.
    \item We conduct ablations to isolate the contributions of the two branches, sensitivity analyses over hyperparameters and an out-of-distribution evaluation to explore behaviour under popularity-distribution shifts.
\end{itemize}

\section{Related Work}

\subsection{Pointwise and Pairwise Objectives}
Pointwise and pairwise losses are the most common objectives used to train implicit-feedback recommender systems \cite{wu2024effectiveness}. Pointwise losses treat the recommendation as a classification or regression task, using losses such as mean squared error (MSE) or binary-cross entropy (BCE) applied to each positive and negative instance separately \cite{Hu2008ICDM_ImplicitMF, He2017WWW_NCF}. Most pointwise formulations take into account instance confidence, weighting observed positives and missing negatives differently \cite{Hu2008ICDM_ImplicitMF, He2017WWW_NCF}. Many deep collaborative models adopt this perspective (observed vs. unobserved), changing the scoring function while keeping the same supervision signal \cite{He2017WWW_NCF}.

Pairwise losses optimise relative preferences among items, attempting to rank positive items above negative items by increasing the gap between their predicted scores, which determines the ordering. Bayesian Personalized Ranking (BPR) formalises this idea via a probabilistic pairwise criterion (often referred to as BPR-Opt) \cite{rendle2012bpr}. 
This pairwise view aligns more clearly than pointwise classification, with the desired ranking behaviour of common recommender systems \cite{rendle2012bpr, wu2024effectiveness}.

\subsection{Softmax-style Objectives}

SL and its scalable approximation sampled SL have garnered more attention recently in recommender systems \cite{wu2024effectiveness}. As the name suggests, SL applies a softmax over predicted scores to obtain a multinomial distribution and maximises the probability of positives relative to negatives. It has a direct link with contrastive objectives, and popular Information Noise-Contrastive Estimation (InfoNCE) can be viewed as a variant of sampled softmax \cite{oord2018CPCrepresentation}. The temperature, $\tau$ governs how much the gradient update is concentrated on the hardest negatives as opposed to being spread evenly across the negatives of all difficulties. We go more into further depth on SL's formulation in Section \ref{sec:SL}. 

In a recent analysis of SL, Wu et al. confirmed the benefits of sampled SL in the realm of item recommendation, attributing those benefits to stronger hard-negative emphasis and closer alignment with ranking quality metrics (like NDCG) \cite{oord2018CPCrepresentation,jarvelin2002cumulated_NDCG}. They also highlight failure modes tied to representation scaling choices, backbone choices and loss interactions. They symmetrise SL robustness to negatives through additional positive regularisation. This extends the log-mean-exponent structure, thus reducing sensitivity to noisy positives \cite{wu2024bsl}. Analysing from a pairwise perspective, Yang et al. introduced PSL, which modifies the surrogate activation function to tighten SL's interpretation as a DCG surrogate loss \cite{yang2024psl}. The choice of activation (ReLU or Tanh) is highlighted as improving robustness by mitigating the impact of false negatives. Finally, to address top-$K$ truncation, Yang et al. propose SoftmaxLoss@K (SL@K), which augments SL with a top-$K$ quantile weighting for each user. This acts as a threshold that separates the top-$K$ positive items from the rest. A Monte Carlo-based quantile estimation strategy is developed to achieve computational efficiency. The SL@K loss can be thought of as extending the idea of SL as a DCG surrogate, to being a DCG@K surrogate, with closer alignment to NDCG@K metrics \cite{yang2025breaking,jarvelin2002cumulated_NDCG}. 

All these variants show a clear trend of treating SL as a flexible backbone loss, and then shaping its behaviour through symmetry, pairwise surrogacy or top-$K$ extensions, addressing the challenges of implicit-feedback recommendation. However, one of the main limitations of these objectives is the assumption of a single global temperature and a uniform treatment of the sampled negatives, even though their quality and competitiveness may vary greatly between different user-positive pairs. Additionally, without impression logs (i.e. when true exposure is unknown), uniform sampling would include items that were never plausible co-exposed competitors \cite{marlin2009collaborative_MNAR}. The presence of such negatives weakens SL's interpretation as a DCM \cite{mcfadden1972conditional}. With this limitation in mind, DSL is designed to reshape the effective temperature from the sampled competition itself, bridging the gap between the true slate-conditioned choice intuition and the reality of training under an implicit-feedback setting.

\section{Preliminaries}
\subsection{Task Formulation}
Our discussion is carried out within the collaborative filtering (CF)~\cite{rendle2012bpr} setting in which supervision is obtained from user--item interaction logs (implicit-feedback). $\mathcal{U}$ and $\mathcal{I}$ denote the user set and item set, respectively. A CF dataset $\mathcal{D}\subseteq \mathcal{U}\times \mathcal{I}$ is a collection of observed interactions, where each instance $(u,i)\in \mathcal{D}$ indicates that user $u$ has interacted with item $i$. For a user $u$, the set of observed (positive) items is defined as
\begin{equation}
\mathcal{P}_u=\{\, i\in \mathcal{I} : (u,i)\in \mathcal{D}\,\},
\end{equation}
while the negatives for training are unobserved items $\mathcal{I}\setminus \mathcal{P}_u$. For each observed pair $(u,i)$, we sample a set of $N$ negatives
\begin{equation}
\mathcal{N}_u \subset \mathcal{I}\setminus \mathcal{P}_u,
\qquad |\mathcal{N}_u|=N,
\end{equation}
where we use $j$ to denote negative items (i.e., $j\in \mathcal{N}_u$).

The objective is to learn a scoring function
$f(u,i):\mathcal{U}\times \mathcal{I}\rightarrow \mathbb{R}$ that quantifies the preference of user $u$ for item $i$. An embedding-based approach is commonly used in modern recommender systems (RS) \cite{koren2009matrix}. Here, we represent each user $u$ and item $i$ using $d$-dimensional embeddings $\mathbf{u}_u,\mathbf{v}_i\in\mathbb{R}^d$. We then compute the preference score from their embedding similarity. We use cosine similarity:
\begin{equation}
f(u,i)=\frac{\mathbf{u}_u^\top \mathbf{v}_i}{\lVert \mathbf{u}_u\rVert_2\, \lVert \mathbf{v}_i\rVert_2}.
\end{equation}
The scores $f(u,i)$ are then used to rank items for generating recommendations.

Additionally, we define the item--item similarity between the positive item $i$ and a negative item $j$ as
\begin{equation}
s_{ij}
= \cos(\mathbf{v}_i,\mathbf{v}_j)
= \frac{\mathbf{v}_i^\top \mathbf{v}_j}
{\lVert \mathbf{v}_i\rVert_2 \, \lVert \mathbf{v}_j\rVert_2}
\label{eq:item-sim}
\end{equation}

\subsection{Softmax Loss}
\label{sec:SL}
The Softmax Loss (SL) has the log-sum-exp form given by \cite{wu2024effectiveness}:
\begin{equation}
\mathcal{L}_{\mathrm{SL}}
= 
\sum_{(u,i)\in\mathcal{B}}
\log \sum_{j\in \mathcal{N}_u}\exp\!\left(\frac{f(u,j)-f(u,i)}{\tau}\right),
\label{eq:softmax-loss}
\end{equation}
where $f(u,i)$ and $f(u,j)$ denote user scores for positive item $i$ and negative item $j$. 

SL is a smooth surrogate for Discounted Cumulative Gain (DCG) \cite{bruch2019analysis}. Let $r_u(i)\in\{1,2,\dots\}$ denote the (1-indexed) ranking position of item $i$ under scores $f(u,\cdot)$.
For binary relevance, the user-level DCG can be written as
\begin{equation}
\mathrm{DCG}(u)
=
\sum_{i\in\mathcal{P}_u}\frac{1}{\log_2(1+r_u(i))},
\label{eq:dcg-def}
\end{equation}
where $\mathcal{P}_u$ is the set of observed positives for user $u$.
A standard relaxation uses $\log_2(1+\pi_u(i)) \le \pi_u(i)$ and Jensen's inequality to obtain
\begin{equation}
-\log \mathrm{DCG}(u) + \log|\mathcal{P}_u|
\;\le\;
\frac{1}{|\mathcal{P}_u|}
\sum_{i\in\mathcal{P}_u}\log r_u(i).
\label{eq:dcg-relax}
\end{equation}
Expressing the rank position through pairwise comparisons and writing the margin
$d_{uij}=f(u,j)-f(u,i)$, we get:
\begin{equation}
r_u(i)
=
1+\sum_{j\in\mathcal{I}\setminus\{i\}}
\mathbb{I}\!\left(f(u,j)\ge f(u,i)\right)
=
1+\sum_{j\in\mathcal{I}\setminus\{i\}} \delta(d_{uij}),
\label{eq:rank-step}
\end{equation}
where $\delta(\cdot)$ is the Heaviside step function.
For any $\tau>0$ we get $\delta(d)\le \exp(d/\tau)$, and we can define the smooth upper bound:
\begin{equation}
\log \pi_u(i)
=
\log\sum_{j\in\mathcal{I}}\delta(d_{uij})
\;\le\;
\log\sum_{j\in\mathcal{I}}\exp\!\left(\frac{d_{uij}}{\tau}\right).
\label{eq:rank-lse-bound}
\end{equation}
Substituting \eqref{eq:rank-lse-bound} into \eqref{eq:dcg-relax} yields that SL (with $\mathcal{I}$
or a sampled approximation of $\mathcal{I}$) serves as a smooth upper bound on the negative log-DCG,
providing a principled surrogate objective for ranking quality (and similarly for mean reciprocal rank (MRR)).\footnote{In practice $\mathcal{I}$ is approximated by $\mathcal{N}_u$ in \eqref{eq:softmax-loss}.}

\section{Methodology}
\subsection{Within Example Weighted Competition ($\kappa$ Branch)}
\label{sec:kappa-branch}
The sampled Softmax Loss (SL) in Eq.~\eqref{eq:softmax-loss} treats all sampled negatives $j\in\mathcal{N}_u$ symmetrically within each training instance $(u,i)$. However, in implicit-feedback collaborative filtering (CF), some negatives are more \emph{competitive} than others---e.g., they may already score highly for $u$ (hard negatives), or be highly similar to the positive item $i$ (near-miss alternatives). To model this heterogeneity without changing the sampler, we introduce a per-negative \emph{competition weight} $\kappa_{uij}>0$ that scales each logit difference inside the softmax.

\subsubsection{Constructing $\kappa_{uij}$ from hardness and item--item similarity:}
The negative score $f(u,j)$ is used as a hardness proxy. High $f(u,j)$ indicates that the model estimates negative $j$ as highly relevant to user $u$ at this point in training. Additionally, item--item similarity between the positive item $i$ and negative item $j$ is used as a proxy for substitutability (nearest-neighbour alternatives). To ensure the similarity range is $s_{ij}\in[0,1]$, avoiding negative $s_{ij}$ for numerical stability, we use the shifted form of Eq.\ref{eq:item-sim}:
\begin{equation}
\bar{s}_{ij} \;=\; \frac{s_{ij}+1}{2}\in[0,1].
\label{eq:shiftedsim}
\end{equation}
High $s_{ij}$ means $j$ and $i$ are alike in representation space, making the negative $j$ a valid alternative that user $u$ could accept. These high $s_{ij}$ are very informative as they represent confusable items. High $s_{ij}$ also suggests that $j$ lies in the same choice set as $i$. Without impression logs, it is a reasonable proxy for co-exposure i.e., the competition that could be modeled with impression logs.

We then form a per-negative \emph{logit} $\ell_{uij}$ by combining hardness and similarity:
\begin{equation}
\ell_{uij}= f(u,j)+\bar{s}_{ij}.
\label{eq:logk}
\end{equation}
We convert $\ell_{uij}$ into positive, mean-one competition weights $\kappa_{uij}$ via a single per-instance
exponential normalization:
\begin{equation}
\kappa_{uij}
=
\frac{\exp(\ell_{uij})}
{\frac{1}{N}\sum_{j'\in\mathcal{N}_u}\exp(\ell_{uij'})},
\qquad\Rightarrow\qquad
\frac{1}{N}\sum_{j\in\mathcal{N}_u}\kappa_{uij}=1.
\label{eq:kappa}
\end{equation}
To control the reweighting strength, we introduce the hyperparameter $\beta\ge 0$:
\begin{equation}
\kappa_{uij}
=
1+\beta\left(\kappa_{uij}-1\right),
\label{eq:kappa-affine}
\end{equation}
which preserves the mean of one property within each example $(u,i)$.

\subsubsection{Per-negative logit scaling:}
With $\kappa_{uij}$ defined, we revisit the softmax loss definition. To aid brevity, we define the score difference for the training triplet $(u,i,j)$ with $j\in\mathcal{N}_u$
\begin{equation}
d_{uij} \;=\; f(u,j)-f(u,i)
\end{equation}
We then augment the softmax loss as $\kappa_{uij}\,d_{uij}$ inside the log-sum-exp, giving the $\kappa$-augmented SL
\begin{equation}
\mathcal{L}_{\kappa\text{-}\mathrm{SL}}
=
\sum_{(u,i)\in\mathcal{B}}
\log
\sum_{j\in\mathcal{N}_u}
\exp\!\left(\frac{\kappa_{uij}\, d_{uij}}{\tau}\right).
\label{eq:kappa-softmax}
\end{equation}
Effectively, a per-negative effective temperature $\tau_{uij}=\tau/\kappa_{uij}$ is induced by $\kappa_{uij}$, such that a bigger $\kappa_{uij}$ sharpens the loss around that negative increasing its gradient contribution and vice versa.

\subsection{Competition-Aware Temperature (CA Branch)}
\label{sec:ca-branch}
The $\kappa$-branch in Sec.~\ref{sec:kappa-branch} models \emph{within-example} heterogeneity by reweighting individual negatives $j\in\mathcal{N}_u$ for a fixed global temperature $\tau$. Complementary to the $\kappa$-branch's within-example reweighting, we introduce a \emph{competition-aware (CA)} temperature that changes \emph{across positive training instances} $(u,i)$. Because uniform sampling does not guarantee an equal number of strong, substitutable negatives, some user--positive pairs are intrinsically more or less ambiguous. 
In general, more competitive instances would benefit from a sharper softmax, whereas less competitive ones would benefit from a smoother softmax for stability.

\subsubsection{Competitor slate:}
For each observed pair $(u,i)$ with $N$ sampled negatives $\mathcal{N}_u$, we form a small competitor slate
\begin{equation}
\mathcal{C}_{ui}\subseteq \mathcal{N}_u, \qquad |\mathcal{C}_{ui}|=K,
\end{equation}
by selecting the top-$K$ negatives with the highest current scores $f(u,j)$. This focuses the CA statistics on the most competitive portion of the sampled set while reducing noise from easy negatives.

\subsubsection{Hardness-induced competitor distribution:}
We convert scores on the competitor slate into a probability distribution over negatives:
\begin{equation}
p_{uij}
=
\frac{\exp\!\left(\frac{f(u,j)}{\tau}\right)}
{\displaystyle\sum_{k\in\mathcal{C}_{ui}}\exp\!\left(\frac{f(u,k)}{\tau}\right)},
\qquad j\in\mathcal{C}_{ui},
\label{eq:ca-pl-prob}
\end{equation}
Intuitively, $p_{uij}$ concentrates mass on the hardest negatives, approximating a local choice set using the model's current scoring mechanism, without impression logs.

\subsubsection{Competition intensity from (hardness $\times$ similarity):}
Using the shifted similarity $\bar{s}_{ij}\in[0,1]$ defined in Eq. \ref{eq:shiftedsim}, we quantify the resemblance of the competitor slate to the positive item under the hardness distribution using the following bounded \emph{log-partition gap} statistic:
\begin{align}
c_{ui}
&=
\log \sum_{j\in\mathcal{C}_{ui}}
\exp\!\left(\frac{f(u,j)}{\tau} + \bar{s}_{ij}\right)
-
\log \sum_{j\in\mathcal{C}_{ui}}
\exp\!\left(\frac{f(u,j)}{\tau}\right)
\label{eq:ca-cui-lse}
\\
&=
\log \,\mathbb{E}_{j\sim p_{ui}}\!\big[\exp(\bar{s}_{ij})\big].
\label{eq:ca-cui-exp}
\end{align}
Because $\bar{s}_{ij}\in[0,1]$, we have $\exp(\bar{s}_{ij})\in[1,e]$ and thus $c_{ui}\in[0,1]$, making the signal well-scaled and stable.

\subsubsection{From competition intensity to per-example temperature:}
We map $c_{ui}$ to a positive \emph{temperature multiplier} $m_{ui}>0$ using exponential tilting followed by mean normalization across the mini-batch $\mathcal{B}$, ensuring the average temperature remains calibrated:
\begin{align}
x_{ui}
&=
\alpha c_{ui}
\label{eq:ca-xui}
\\
m_{ui}
&=
\frac{\exp(x_{ui})}{\frac{1}{|\mathcal{B}|}\displaystyle\sum_{(u',i')\in\mathcal{B}}\exp(x_{u'i'})},
\quad\Rightarrow\quad
\frac{1}{|\mathcal{B}|}\sum_{(u,i)\in\mathcal{B}} m_{ui}=1,
\label{eq:ca-temp-mult}
\end{align}
where $\alpha\ge 0$ controls the modulation strength.
The mean normalization in Eq.~\eqref{eq:ca-temp-mult} is designed to prevent global temperature drift, ensuring results remain comparable between techniques. 

We then define a \emph{per-example} temperature
\begin{equation}
\tau_{ui}=\frac{\tau}{m_{ui}},
\qquad\text{equivalently}\qquad
\frac{1}{\tau_{ui}}=\frac{m_{ui}}{\tau}.
\label{eq:ca-tau-ui}
\end{equation}
Instances with higher competition intensity (larger $c_{ui}$) tend to yield larger $m_{ui}$ and therefore smaller $\tau_{ui}$, sharpening the softmax around these competitive examples.

\subsubsection{CA-softmax objective:}
Let $d_{uij}=f(u,j)-f(u,i)$ as in Sec.~\ref{sec:kappa-branch}. With CA only (i.e., $\kappa_{uij}\equiv 1$), we obtain
\begin{equation}
\mathcal{L}_{\mathrm{CA}\text{-}\mathrm{SL}}
=
\sum_{(u,i)\in\mathcal{B}}
\log
\sum_{j\in\mathcal{N}_u}
\exp\!\left(\frac{d_{uij}}{\tau_{ui}}\right).
\label{eq:ca-softmax}
\end{equation}

\subsection{Combined DSL Objective}
\label{sec:combinedDSL}
When combining both the $\kappa$ and CA branches, the per-negative logit scaling and the per-example temperature act multiplicatively:
\begin{equation}
\mathcal{L}_{\mathrm{DSL}}
=
\sum_{(u,i)\in\mathcal{B}}
\log
\sum_{j\in\mathcal{N}_u}
\exp\!\left(\frac{\kappa_{uij}\,d_{uij}}{\tau_{ui}}\right),
\label{eq:kappa-ca-softmax}
\end{equation}
which corresponds to an effective per-negative temperature $\tau_{uij}=\tau_{ui}/\kappa_{uij}$.

\subsubsection{Final $\tau$ drift control:}
Although the row-wise mean-one constraint $\mathbb{E}_{j\in\mathcal{N}_u}[\kappa_{uij}]=1$ is enforced, this does not control the average inverse weight, and in general $\mathbb{E}_{j}[1/\kappa_{uij}]\neq 1$. By Jensen's inequality, for non-constant $\kappa_{uij}>0$ we have
$\mathbb{E}[1/\kappa_{uij}] \ge 1/\mathbb{E}[\kappa_{uij}]=1$.
We therefore rescale by $\mathbb{E}_{j}[1/\kappa_{uij}]$ to prevent unintended logit-scale (temperature) drift when combining $\kappa$ with CA.

\section{DSL Theoretical Analysis}
\subsection{Variational view of sampled Softmax and KL-DRO}
\label{sec:theory-variational-sl}

The standard change-of-measure (Donsker--Varadhan / Gibbs) variational identity connects log-sum-exp to KL-based Distributionally Robust Optimization (KL-DRO).
\cite{donsker1975asymptotic,namkoong2016stochastic}.

The uniform reference distribution over negatives is $\pi_u(j)=1/N$ for $j\in\mathcal{N}_u$. The (normalised) free-energy form is more convenient to work with for analysis purposes:
\begin{equation}
\tilde{\ell}^{\mathrm{SL}}_{ui}(\tau)
\;=\;
\tau \log \Bigg(
\mathbb{E}_{j\sim \pi_u}\Big[\exp\!\big(\tfrac{d_{uij}}{\tau}\big)\Big]
\Bigg)
\;=\;
\tau \log \Bigg(
\frac{1}{N}\sum_{j\in\mathcal{N}_u}\exp\!\big(\tfrac{d_{uij}}{\tau}\big)
\Bigg),
\label{eq:sl-free-energy}
\end{equation}
which differs from the sampled Softmax loss ($\log\sum_{j\in\mathcal{N}_u}\exp(d_{uij}/\tau)$)
only by the constant $\tau\log N$.

The variational identity states that for any base measure $\pi$ and any measurable $h$,
\begin{equation}
\log \,\mathbb{E}_{j\sim \pi}\!\big[\exp(h(j))\big]
\;=\;
\sup_{q\in\Delta}\Big\{\mathbb{E}_{j\sim q}[h(j)] - D_{\mathrm{KL}}(q\,\|\,\pi)\Big\},
\label{eq:dv-identity}
\end{equation}
where $\Delta$ is the probability simplex and $D_{\mathrm{KL}}(\cdot\|\cdot)$ is the Kullback--Leibler (KL) divergence.

Applying \eqref{eq:dv-identity} to $h(j)=d_{uij}/\tau$ with $\pi=\pi_u$ yields the KL-DRO form of sampled Softmax:
\begin{equation}
\tilde{\ell}^{\mathrm{SL}}_{ui}(\tau)
\;=\;
\sup_{q_{ui}\in \Delta(\mathcal{N}_u)}
\Big\{
\mathbb{E}_{j\sim q_{ui}}[d_{uij}]
\;-\;
\tau\,D_{\mathrm{KL}}(q_{ui}\,\|\,\pi_u)
\Big\}.
\label{eq:sl-kldro}
\end{equation}
The optimiser is the Gibbs/softmax reweighting over negatives,
\begin{equation}
q^{\star}_{uij}(\tau)
\;=\;
\frac{\pi_u(j)\exp\!\big(\tfrac{d_{uij}}{\tau}\big)}
{\mathbb{E}_{k\sim\pi_u}\big[\exp\!\big(\tfrac{d_{uik}}{\tau}\big)\big]}
\;=\;
\frac{\exp\!\big(\tfrac{d_{uij}}{\tau}\big)}
{\sum_{k\in\mathcal{N}_u}\exp\!\big(\tfrac{d_{uik}}{\tau}\big)}.
\label{eq:sl-qstar}
\end{equation}

Equation \eqref{eq:sl-kldro} demonstrates the direct role of temperature, such that $\tau$ is the penalty weight on deviating from the reference $\pi_u$. Smaller $\tau$ permits more deviation (a sharper $q^\star$ concentrating on the hardest negatives), while larger $\tau$ enforces $q^\star$ being closer to $\pi_u$, which is a smoother and less robust reweighting.
This is also seen from the classical log-sum-exp smoothing bound \cite{BoydVandenberghe2004}:
\begin{equation}
\max_{j\in\mathcal{N}_u} d_{uij} - \tau\log N
\;\le\;
\tilde{\ell}^{\mathrm{SL}}_{ui}(\tau)
\;\le\;
\max_{j\in\mathcal{N}_u} d_{uij},
\label{eq:sl-smoothmax}
\end{equation}
so the approximation gap to the hard maximum is controlled by $\tau\log N$.

\subsection{Implications of $\kappa$ and CA for robustness (KL-DRO view)}
\label{sec:theory-dsl-robustness}

Let $\pi_u$ be uniform on $\mathcal{N}_u$ ($\pi_u(j)=1/N$).
The DSL payoff is defined as $a_{uij}=\kappa_{uij}\,d_{uij}$ and the CA temperature is $\tau_{ui}=\tau/m_{ui}$.
Unlike Eq.~\eqref{eq:kappa-ca-softmax}, this expression includes an extra term $\tau_{ui}\log N$
\begin{equation}
\tilde{\ell}^{\mathrm{DSL}}_{ui}
\!=\!
\tau_{ui}\log\Bigg(\mathbb{E}_{j\sim\pi_u}\Big[\exp\!\Big(\frac{a_{uij}}{\tau_{ui}}\Big)\Big]\Bigg)
\!=\!
\tau_{ui}\log\Bigg(\frac{1}{N}\sum_{j\in\mathcal{N}_u}\exp\!\Big(\frac{\kappa_{uij}d_{uij}}{\tau_{ui}}\Big)\Bigg).
\label{eq:theory-dsl-free-energy}
\end{equation}
Using the standard Gibbs / Donsker--Varadhan variational identity, $\tilde{\ell}^{\mathrm{DSL}}_{ui}$ yields the KL-DRO form \cite{donsker1975asymptotic}:
\begin{equation}
\tilde{\ell}^{\mathrm{DSL}}_{ui}
\;=\;
\sup_{q_{ui}\in\Delta(\mathcal{N}_u)}
\Big\{
\mathbb{E}_{j\sim q_{ui}}[a_{uij}]
-
\tau_{ui}\,D_{\mathrm{KL}}(q_{ui}\,\|\,\pi_u)
\Big\},
\label{eq:theory-dsl-kldro}
\end{equation}
with optimizer
\begin{equation}
q^{\star}_{uij}
=
\frac{\exp\!\big(a_{uij}/\tau_{ui}\big)}
{\sum_{k\in\mathcal{N}_u}\exp\!\big(a_{uik}/\tau_{ui}\big)}
=
\frac{\exp\!\big(\kappa_{uij}d_{uij}/\tau_{ui}\big)}
{\sum_{k\in\mathcal{N}_u}\exp\!\big(\kappa_{uik}d_{uik}/\tau_{ui}\big)}.
\label{eq:theory-dsl-qstar}
\end{equation}

Compared to standard SL Eq.~\eqref{eq:theory-dsl-kldro} exhibits two direct robustness effects: (i) $\kappa$ changes the robustness payoff from
$d_{uij}$ to $\kappa_{uij}d_{uij}$ (so the ``worst-case negative'' is taken over \emph{rescaled} margins), and
(ii) CA changes the robustness level per example via the KL penalty weight $\tau_{ui}$.
A compact worst-case bound follows from log-sum-exp smoothing:
\begin{equation}
\max_{j\in\mathcal{N}_u} a_{uij} \;-\; \tau_{ui}\log N
\;\le\;
\tilde{\ell}^{\mathrm{DSL}}_{ui}
\;\le\;
\max_{j\in\mathcal{N}_u} a_{uij},
\label{eq:theory-dsl-smoothmax}
\end{equation}
so smaller $\tau_{ui}$ makes the objective more ``max-like'' over $\kappa_{uij}d_{uij}$ (more robust / more peaked $q^\star$),
while larger $\tau_{ui}$ enforces $q^\star$ being closer to uniform.

Let $\lambda_{ui}=1/\tau_{ui}$ and define the log-partition
\begin{equation}
A_{ui}(\lambda)
=
\log\mathbb{E}_{j\sim\pi_u}\big[\exp(\lambda a_{uij})\big].
\label{eq:theory-Aui}
\end{equation}
The KL radius actually attained by the adversary is
\begin{equation}
\rho_{ui}(\lambda)
:=
D_{\mathrm{KL}}(q^\star_{ui}(\lambda)\,\|\,\pi_u)
=
\lambda A'_{ui}(\lambda)-A_{ui}(\lambda),
\label{eq:theory-rho}
\end{equation}
and it satisfies the monotonicity \cite{wainwright2008graphical}
\begin{equation}
\frac{d}{d\lambda}\rho_{ui}(\lambda)
=
\lambda A''_{ui}(\lambda)
=
\lambda\,\mathrm{Var}_{j\sim q^\star_{ui}(\lambda)}[a_{uij}]
\;\ge\;0.
\label{eq:theory-rho-monotone}
\end{equation}
Hence, when CA produces larger $m_{ui}$ (so larger $\lambda_{ui}=m_{ui}/\tau$) and $\kappa$ assigns larger weights for harder and semantically close negatives, the adversary provably deviates further from uniform in KL, strengthening the KL-DRO robustness for that example. Robustness is therefore adaptively concentrated where it matters more.
This leads us to the guarantee that CA increases the attained KL deviation when it sharpens.

A limitation that arises from this analysis is that, because sharpening concentrates mass on the tail of the sampled negatives, DSL will be more sensitive to false negatives (noise). Intuitively, because DRO assigns larger losses to harder instances, it will amplify the contribution of noisy data for optimisation. This has been demonstrated in many DRO studies \cite{zhai2021doro,liu2023geometry}.

\subsubsection{Calibration and drift:}
Because $\frac{1}{|\mathcal{B}|}\sum_{(u,i)\in\mathcal{B}} m_{ui}=1$ (Eq.~\eqref{eq:ca-temp-mult}), CA preserves the batch-average inverse temperature:
\begin{equation}
\frac{1}{|\mathcal{B}|}\sum_{(u,i)\in\mathcal{B}}\frac{1}{\tau_{ui}}
=
\frac{1}{|\mathcal{B}|}\sum_{(u,i)\in\mathcal{B}}\frac{m_{ui}}{\tau}
=
\frac{1}{\tau}.
\label{eq:theory-ca-calibration}
\end{equation}
When combining with $\kappa$, the effective per-negative temperature is $\tau_{uij}=\tau_{ui}/\kappa_{uij}$, so even with $\mathbb{E}_j[\kappa_{uij}]=1$,
\begin{equation}
\mathbb{E}_{j\sim\pi_u}[\tau_{uij}]
=
\tau_{ui}\,\mathbb{E}_{j\sim\pi_u}\!\Big[\frac{1}{\kappa_{uij}}\Big]
\;\ge\;
\tau_{ui},
\label{eq:theory-invmean-drift}
\end{equation}
motivating the inverse-mean drift control described in Section \ref{sec:combinedDSL}.
Thus, the full technique avoids temperature drift, and results are comparable to those of other approaches using the same base temperature.

\subsection{Metric-Aligned Gradient Estimation of DSL}
\label{sec:metric-alignment}

In metric-driven learning-to-rank, LambdaLoss-style methods weight pairwise updates by the
change in the target metric induced by swapping the ranks of $(i,j)$, i.e., $\Delta\mathrm{NDCG@}K_{uij}$ \cite{jarvelin2002cumulated_NDCG,burges2006learning}.
DSL can be viewed as a sampled-softmax instantiation of this principle: the weight
$\lambda^{\mathrm{DSL}}_{uij}$ is a smooth, readily computable proxy that (i) increases with the likelihood
that $j$ is a Top-$K$ rank threat (via $q_{uij}$), and (ii) increases for semantically substitutable competitors
(via $\kappa_{uij}$), which are the swaps that tend to incur the largest $\Delta\mathrm{NDCG@}K$.
When combined with the Top-$K$ emphasis provided by SL@$K$ (where $g(\mathrm{rank}_K(i))$) DSL concentrates gradient budget on the pairs with the highest expected impact on the Top-$K$ metrics \cite{yang2024psl,yang2025breaking}. This improves the efficiency and stability of the stochastic metric surrogate, when compared to treating sampled negatives symmetrically.

DSL keeps the same log-sum-exp backbone as SL, but modifies the
\emph{competition geometry} in two complementary ways: firstly through \emph{competition-shaped} per-negative
scaling ($\kappa$ branch), and secondly through \emph{competition-aware} per-example temperature modulation (CA branch).
Concretely, for a training pair $(u,i)$ and sampled negatives $\mathcal{N}_u$, we define the margin
$d_{uij}=f(u,j)-f(u,i)$ and the DSL exponent
\begin{equation}
z_{uij} \;=\; \frac{\kappa_{uij}\,d_{uij}}{\tau_{ui}},
\qquad
q_{uij}\;=\;\frac{\exp(z_{uij})}{\sum_{k\in\mathcal{N}_u}\exp(z_{uik})}.
\label{eq:dsl-q}
\end{equation}
For the per-positive loss $\ell_{ui}=\log\sum_{j\in\mathcal{N}_u}\exp(z_{uij})$, the score gradients are
\begin{equation}
\frac{\partial \ell_{ui}}{\partial f(u,j)}
=
q_{uij}\,\frac{\kappa_{uij}}{\tau_{ui}},
\qquad
\frac{\partial \ell_{ui}}{\partial f(u,i)}
=
-\sum_{j\in\mathcal{N}_u} q_{uij}\,\frac{\kappa_{uij}}{\tau_{ui}}.
\label{eq:dsl-grads}
\end{equation}
Thus, DSL induces an implicit \emph{pairwise update weight}
\begin{equation}
\lambda^{\mathrm{DSL}}_{uij}
\;\triangleq\;
q_{uij}\,\frac{\kappa_{uij}}{\tau_{ui}}.
\label{eq:dsl-lambda}
\end{equation}

The softmax factor $q_{uij}$ is monotonically increasing in the margin $d_{uij}$, so it concentrates mass on
the highest-scoring negatives. These are the items most likely to swap
order with $i$ inside the Top-$K$ band, and therefore most likely to affect the ranking and change the Recall@$K$ and NDCG@$K$.

\subsubsection{Within-example competition shaping ($\kappa$ branch):}
The per-negative multiplier $\kappa_{uij}$ further redistributes update mass \emph{within} each sampled set
$\mathcal{N}_u$ without changing the sampler.
It increases the effective slope of the comparisons deemed most \emph{competitive} (e.g., near-substitutes or
highly confusable rivals), while preserving the log-sum-exp structure.
Operationally, $\kappa_{uij}$ scales both the exponent that defines $q_{ui}$ and the resulting gradient
weight in \eqref{eq:dsl-grads}, so confusable negatives receive a larger share of the stochastic gradient exactly
where pairwise swaps are most impactful for Top-$K$ metrics.

\subsubsection{Across-example competition awareness (CA branch):}
Uniform sampling yields heterogeneous training instances where some $(u,i)$ pairs have stronger, more substitutable
sampled negatives in $\mathcal{N}_u$ than others. The CA branch addresses this \emph{across positives} by
constructing a small competitor slate $\mathcal{C}_{ui}$ (top-$K$ by $f(u,j)$), estimating a bounded competition
intensity $c_{ui}$ on that slate, and mapping it to a multiplier $m_{ui}$ with mean normalization.
The resulting per-example temperature $\tau_{ui}=\tau/m_{ui}$ sharpens the softmax for intrinsically ambiguous
instances (large $c_{ui}$) and smooths it otherwise, stabilising learning while increasing gradient concentration on examples where Top-$K$ errors are most likely.

\section{Experiments}
We first conduct experiments to demonstrate the performance of DSL and DSL@K compared to SL, SL@K and other state-of-the-art losses. We then dig deeper into the properties of DSL, ablating the contributions of the $\kappa$ and $CA$ branches. We explore DSL's robustness to out-of-distribution (OOD) data. Finally, we compare SL and DSL performance with head and tail items.

\subsection{Experimental Setup}
\subsubsection{Datasets:} To enable fair comparisons, our experimental protocol is consistent with Yang et al. \cite{yang2024psl,yang2025breaking} and Wu et al. \cite{wu2024bsl}. We evaluate on four commonly used and publicly available datasets, namely Amazon-Health (Health), Amazon-Electronic (Electronic), Amazon-Movie (Movie) and Gowalla. These datasets are used to evaluate under IID settings (where the distribution of randomly split training and test data is identical). We also evaluate under OOD settings (where the item popularity distribution shifts between training and test sets). Previous work excludes Health and Movie from OOD comparisons because item popularity is not heavily skewed on these datasets, so we only consider Electronic and Gowalla here \cite{yang2024psl}. The datasets are split into 80\% training and 20\% test sets. Under the IID setting, we further split the training test set into 90\% and 10\% validation sets for hyperparameter tuning. To avoid test set information leakage, no validation set is introduced under the OOD setting. 

\subsubsection{Metrics:}
Similar to Wu et al. and Zhang et al., NDCG@K and Recall@K are utilised for performance evaluation and comparison. The neighbourhood size is set to $K=20$ as in recent works \cite{zhang2023empowering,wu2024bsl}. However, similar results were obtained with different $K$ choices.

\subsubsection{Recommendation backbones:} 
DSL and all baseline losses are evaluated on two representative backbones that are central to modern recommender systems, MF and LightGCN.

\subsubsection{Baseline Losses:}
We compare against seven loss functions. These include the commonly adopted pairwise loss BPR, Partial AUC surrogate loss LLPAUC, NDCG surrogate losses, GuidedRec and SL, and the more recent SL enhancements, AdvInfoNCE, BSL and PSL.  

\subsubsection{Hyperparameter Settings:}
The Adam optimiser is used for all training. The learning rate is varied in $\{10^{-1},10^{-2},10^{-3}\}$, except for BPR where $10^{-4}$ is used as recommended. The weight decay factor (wd) is searched over $\{0,10^{-4},10^{-5},10^{-6}\}$. Batch size is set to $1024$, and we train for $200$ epochs. Following previous work \cite{wu2024bsl,yang2024psl,yang2025breaking}, $1000$ negative items per positive are uniformly sampled.

Interestingly, during our analysis it became apparent that the gains presented in some prior works appear inflated. To resolve this, we conduct an extensive grid search over the hyperparameter configurations of each technique, determining their individually optimal hyperparameters. See Appendix \ref{appendix} for the optimal configurations used for our analysis.

We tune the loss-specific hyperparameters as follows.
\begin{itemize}
  \item \textbf{BPR:} no additional hyperparameters \cite{rendle2012bpr}.
  \item \textbf{LLPAUC:} following Shi et al.~\cite{shi2024lower}, we search
  $\alpha$ in \{0.1, 0.3, 0.5, 0.7, 0.9\} and $\beta$ $\{0.01,0.1\}$.
  \item \textbf{Softmax Loss (SL):} we search
  $\tau$ in \{0.005, 0.025, 0.05, 0.1, 0.2, 0.25\} \cite{yang2024psl}.
  \item \textbf{AdvInfoNCE:} we search $\tau$ same as SL, and fix the remaining
  hyperparameters to the defaults of Zhang et al.~\cite{}. In particular, we set the
  negative weight to $64$, perform adversarial updates every $5$ epochs, and use an adversarial
  learning rate of $5 \times 10^{-5}$ \cite{zhang2023empowering}.
  \item \textbf{BSL:} the positive- and negative-term temperatures $(\tau_1,\tau_2)$
  are searched independently using the same candidate set as SL \cite{wu2024bsl}.
  \item \textbf{PSL:} we tune $\tau$ over the same grid as SL \cite{yang2024psl}.
  \item \textbf{SL@K:} $\tau$ is set to best SL $\tau$, $\tau_w \in \{0.5,3.0\}$ with 0.25 search step and $T_\beta$ in \{5, 20\} \cite{yang2025breaking}.
  \item For \textbf{DSL:} $\tau$ is searched same as SL and  $\beta$ and $\alpha$ are searched in $\{1.0,1.5,2.0,2.5,3.0\}$. We set $C_{ui}=20$.
\end{itemize}

\subsection{Performance Comparison}
Table \ref{tab:iid-main} demonstrates how the performance of DSL compares to SL and other baseline losses on MF and LightGCN backbones. These results illustrate substantial improvements with DSL over other baselines. When compared to SL, in particular, gains on Recall@20 and NDCG@20 can exceed 10\%. On average, across all datasets, metrics and backbones, DSL outperforms SL by an average of \textbf{6.22\%}. It also surpasses all other baselines, indicating that gains are not backbone or dataset-specific, but reflect a more effective objective.

These results align with our metric-driven analysis of DSL in Section \ref{sec:metric-alignment}. By applying pairwise weights and adjusting per-instance $\tau_{ui}$, DSL allows gradients to concentrate on hard and highly similar negatives that are likely to incur the largest $\Delta\mathrm{NDCG@}K$ and $\Delta\mathrm{Recall@}K$ when swapped. While maintaining the log-sum-exp form, DSL reshapes competition within and across training examples, resulting in the consistent Top-$K$ improvements observed in Table \ref{tab:iid-main}.

\begin{table*}[t]
\centering
\small
\setlength{\tabcolsep}{4pt}
\renewcommand{\arraystretch}{1.15}
\begin{tabular}{c|l|cc|cc|cc|cc}
\hline
\multirow{2}{*}{\textbf{Backbone}} & \multirow{2}{*}{\textbf{Loss}}
& \multicolumn{2}{c|}{\textbf{Health}} & \multicolumn{2}{c|}{\textbf{Electronic}}
& \multicolumn{2}{c|}{\textbf{Movie}}  & \multicolumn{2}{c}{\textbf{Gowalla}} \\
& & \textbf{Recall@20} & \textbf{NDCG@20} & \textbf{Recall@20} & \textbf{NDCG@20} & \textbf{Recall@20} & \textbf{NDCG@20} & \textbf{Recall@20} & \textbf{NDCG@20} \\
\hline
\multirow{10}{*}{MF} & BPR & 0.1240 & 0.0815 & 0.0406 & 0.0236 & 0.0428 & 0.0272 & 0.0554 & 0.0382 \\
 & GuidedRec & 0.1478 & 0.1002 & 0.0645 & 0.0383 & 0.0612 & 0.0405 & 0.1129 & 0.0853 \\
 & LLPAUC & 0.1627 & 0.1185 & 0.0829 & 0.0501 & 0.1264 & 0.0880 & 0.1611 & 0.1184 \\
 & SL & 0.1749 & 0.1261 & 0.0825 & 0.0531 & 0.1285 & 0.0924 & 0.2069 & 0.1623 \\
 & AdvInfoNCE & 0.1757 & 0.1282 & 0.0826 & 0.0527 & 0.0994 & 0.0721 & 0.2069 & 0.1625 \\
 & BSL & 0.1749 & 0.1261 & 0.0829 & 0.0527 & 0.1004 & 0.0742 & 0.2068 & 0.1623 \\
 & PSL & 0.1713 & 0.1270 & 0.0833 & 0.0537 & \underline{0.1296} & \underline{0.0935} & 0.2090 & 0.1645 \\
 & SL@K & \underline{0.1756} & \textbf{0.1355} & \textbf{0.0893} & \textbf{0.0585} & 0.1155 & 0.0832 & \textbf{0.2128} & \textbf{0.1709} \\
 & \textbf{DSL (Ours)} & \textbf{0.1871} & \underline{0.1350} & \underline{0.0884} & \underline{0.0577} & \textbf{0.1336} & \textbf{0.0954} & \underline{0.2099} & \underline{0.1705} \\
\cline{2-10}
 & \textcolor{red}{Imp.\%} & \textcolor{red}{+6.98\%} & \textcolor{red}{+7.06\%} & \textcolor{red}{+7.15\%} & \textcolor{red}{+8.66\%} & \textcolor{red}{+3.97\%} & \textcolor{red}{+3.25\%} & \textcolor{red}{+1.45\%} & \textcolor{red}{+5.05\%} \\
\hline
\multirow{10}{*}{LightGCN} & BPR & 0.1177 & 0.0759 & 0.0297 & 0.0173 & 0.0488 & 0.0314 & 0.0433 & 0.0283 \\
 & GuidedRec & 0.1358 & 0.0892 & 0.0614 & 0.0364 & 0.0594 & 0.0392 & 0.0766 & 0.0533 \\
 & LLPAUC & 0.1658 & 0.1173 & 0.0823 & 0.0501 & 0.1263 & 0.0879 & 0.1551 & 0.1134 \\
 & SL & 0.1703 & 0.1230 & 0.0774 & 0.0485 & \underline{0.1295} & 0.0930 & 0.2006 & 0.1549 \\
 & AdvInfoNCE & 0.1704 & 0.1217 & 0.0759 & 0.0478 & 0.0998 & 0.0738 & 0.2001 & 0.1544 \\
 & BSL & 0.1703 & 0.1230 & 0.0774 & 0.0485 & 0.1023 & 0.0758 & 0.2005 & 0.1549 \\
 & PSL & 0.1619 & 0.1184 & 0.0766 & 0.0483 & 0.1290 & \underline{0.0940} & \underline{0.2026} & \underline{0.1570} \\
 & SL@K & \underline{0.1723} & \underline{0.1311} & \underline{0.0842} & \textbf{0.0539} & 0.1069 & 0.0812 & 0.2004 & 0.1564 \\
 & \textbf{DSL (Ours)} & \textbf{0.1875} & \textbf{0.1357} & \textbf{0.0850} & \underline{0.0535} & \textbf{0.1328} & \textbf{0.0951} & \textbf{0.2081} & \textbf{0.1654} \\
\cline{2-10}
 & \textcolor{red}{Imp.\%} & \textcolor{red}{+10.10\%} & \textcolor{red}{+10.33\%} & \textcolor{red}{+9.82\%} & \textcolor{red}{+10.31\%} & \textcolor{red}{+2.56\%} & \textcolor{red}{+2.26\%} & \textcolor{red}{+3.74\%} & \textcolor{red}{+6.78\%} \\
\hline
\end{tabular}
\caption{IID performance (Recall@20 and NDCG@20). The best loss per dataset per metric is in bold, and the second best is underlined. \textcolor{red}{Imp.\%} is computed against \textbf{SL}.}
\label{tab:iid-main}
\vspace{-5pt}
\end{table*}

\begin{table}[t]
\centering
\small
\setlength{\tabcolsep}{4pt}
\renewcommand{\arraystretch}{1.15}

\resizebox{\columnwidth}{!}{%
\begin{tabular}{c|l|cc|cc}
\hline
\multirow{2}{*}{\textbf{Backbone}} & \multirow{2}{*}{\textbf{Loss}}
& \multicolumn{2}{c|}{\textbf{Electronic}}
& \multicolumn{2}{c}{\textbf{Gowalla}} \\
& & \textbf{Recall@20} & \textbf{NDCG@20} & \textbf{Recall@20} & \textbf{NDCG@20} \\
\hline
\multirow{10}{*}{MF}
& BPR        & 0.0102 & 0.0057 & 0.0256 & 0.0174 \\
& GuidedRec  & 0.0114 & 0.0063 & 0.0296 & 0.0214 \\
& LLPAUC     & 0.0227 & 0.0138 & 0.0727 & 0.0522 \\
& SL         & 0.0217 & 0.0129 & 0.0972 & 0.0705 \\
& AdvInfoNCE & 0.0191 & 0.0115 & 0.0649 & 0.0463 \\
& BSL        & 0.0228 & 0.0130 & 0.0975 & 0.0709 \\
& PSL        & \underline{0.0237} & \underline{0.0146} & \underline{0.1005} & \underline{0.0743} \\
& SL@K       & 0.0211 & 0.0128 & 0.0980 & 0.0734 \\
& \textbf{DSL (Ours)} & \textbf{0.0240} & \textbf{0.0150} & \textbf{0.1015} & \textbf{0.0747} \\
\cline{2-6}
& \textcolor{red}{Imp.\%} & \textcolor{red}{+10.60\%} & \textcolor{red}{+16.28\%} & \textcolor{red}{+4.42\%} & \textcolor{red}{+5.96\%} \\
\hline
\end{tabular}%
}

\caption{Out-of-distribution (OOD) performance (Recall@20 and NDCG@20). Best loss per metric is in bold, and second best is underlined. \textcolor{red}{Imp.\%} is computed against \textbf{SL} (Softmax).}
\vspace{-10pt}
\label{tab:ood-mf}
\end{table}

\subsection{OOD Performance}

Table \ref{tab:ood-mf} demonstrates performance under the OOD popularity shift setting. Since the trends are consistent across backbones, only the MF results are reported here for brevity. DSL exhibits high robustness to distributional shifts, as shown in the results. An average improvement of 9.31\% over SL is achieved across the Electronic and Gowalla datasets. DSL outperforms all other baselines, too. We also note that gains are larger than on IID setting. For example, Recall@20 on Gowalla increases from +1.45\% to +4.42\% and on Electronic from +7.15\% to +10.60\%. Similar relative increases can also be seen with NDCG@20.

The pattern observed here is consistent with the KL-DRO analysis in Section \ref{sec:theory-dsl-robustness}. DSL still corresponds to a KL-regularized worst-case objective over negatives (as does SL), but the additional $\kappa$ changes the robustness payoff, focusing adversarial learning on strong, highly substitutable competitors. Meanwhile, CA adapts per-example robustness level via $\tau_{ui}$
for ambiguous, shift-prone instances, which is relevant for this popularity shift evaluation. Thanks to the normalisation measures introduced to prevent global temperature drift, these OOD gains can be directly attributed to DSL’s competition shaping.

\subsection{Ablation Study}
\begin{figure}[htbp]
\centering
\includegraphics[width=\columnwidth]{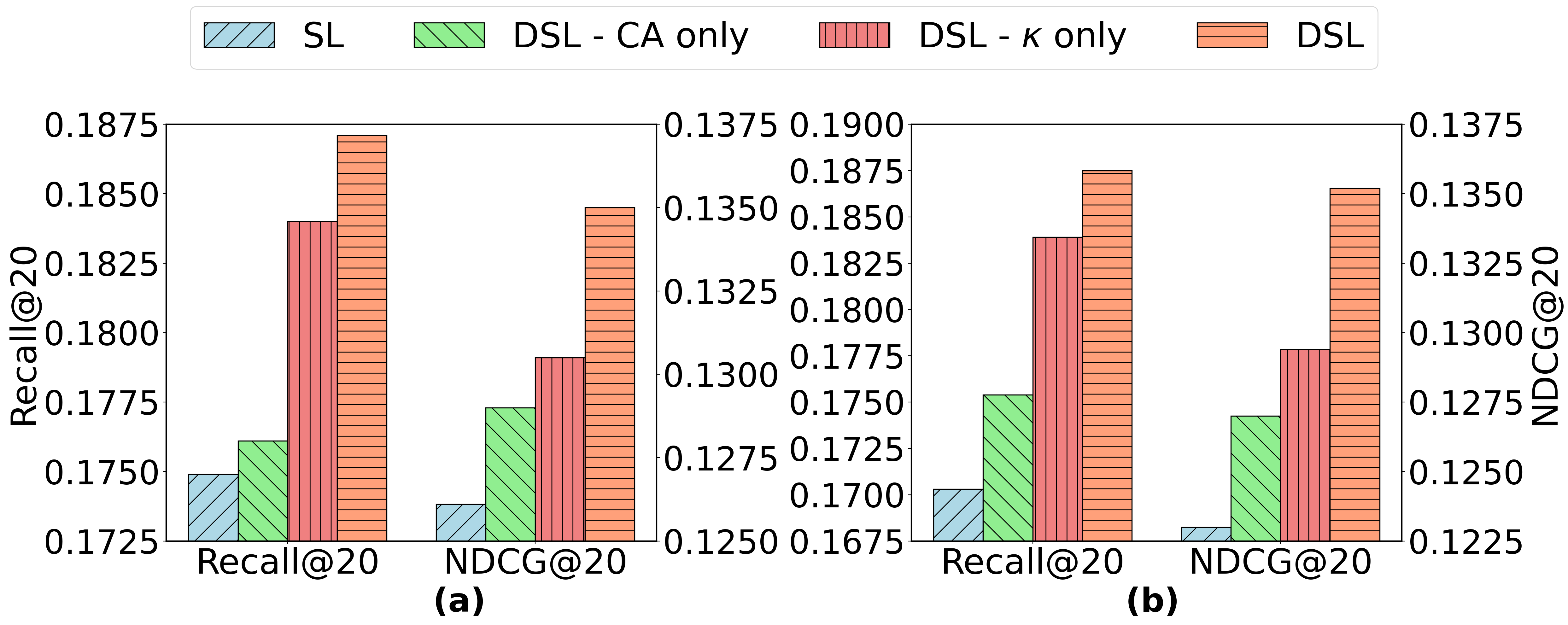}
\includegraphics[width=\columnwidth]{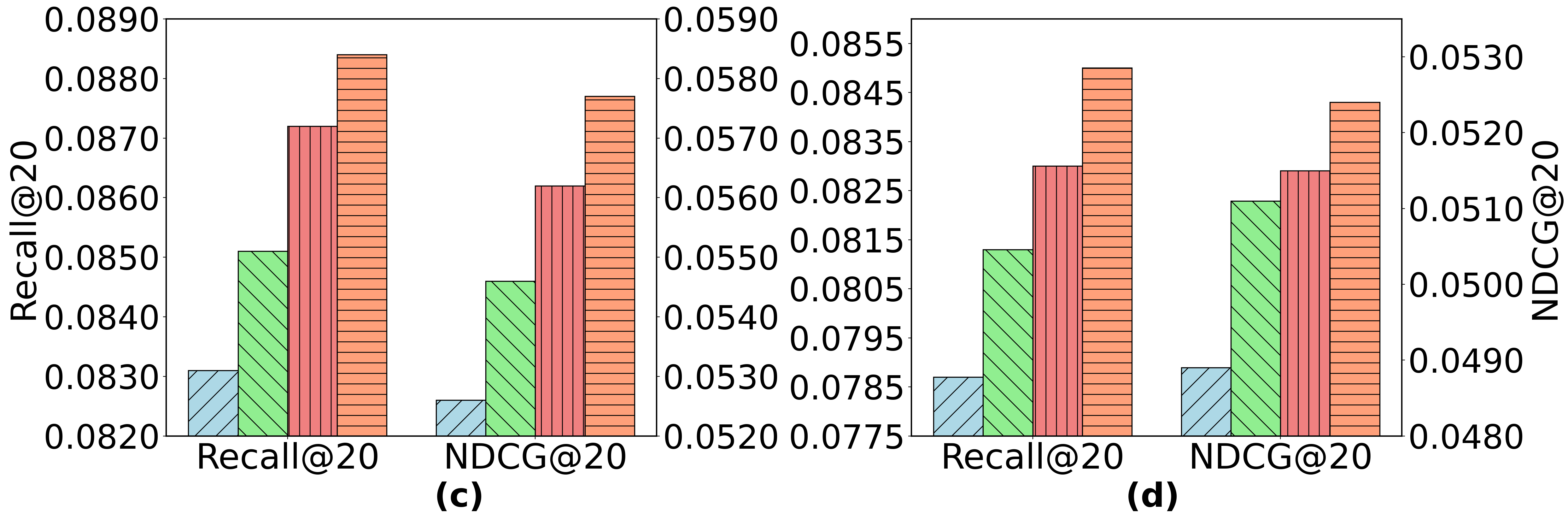}
\caption {Health and Electronic Ablation}
\label{fig-health_elec_ablation}
\end{figure}

We conduct an ablation study to determine the $\kappa$ and $CA$ branches' contributions, separately, and how they influence the combined DSL objective. Figure \ref{fig-health_elec_ablation} reports results on the Health and Electronic datasets on both MF and LightGCN backbones. The four variants compared include (i) \textbf{SL} with both components disabled, (ii) \textbf{DSL--CA only} with the per-instance temperature multiplier enabled only, (iii) \textbf{DSL--$\kappa$ only} with only the per-negative $\kappa$ modulation enabled, and (iv) \textbf{DSL} having both branches enabled. We can observe that enabling either component improves over the base SL, confirming that each branch contributes a useful signal beyond SL's standard log-sum-exponent optimisation. We note that \textbf{DSL--$\kappa$ only} consistently yields bigger improvements than \textbf{DSL--CA only}, suggesting that per-negative reweighting is the stronger driver of gains. Finally, \textbf{full DSL} achieves the best results in all settings, highlighting the clear complementary role that the two branches play. 

\begin{figure}[htbp]
\centering
\includegraphics[width=\columnwidth]{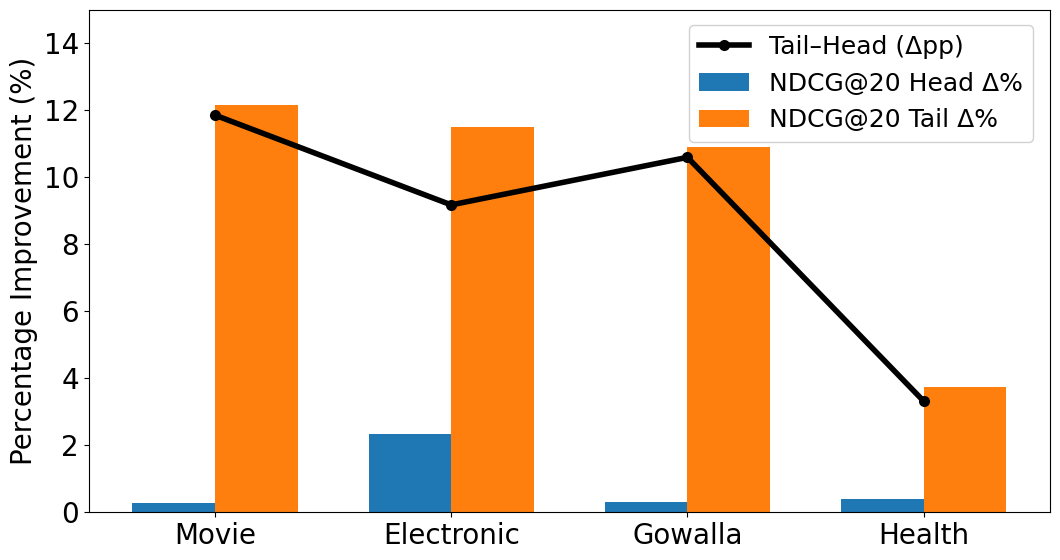}
\caption {Percentage improvements for DSL over SL on NDCG@20 for Head and Tail item buckets, and the Tail--Head gap}
\vspace{-5pt}
\label{fig-head_tail}
\end{figure}

\subsection{Tail and Head Performance Comparison}

Finally, we explore the robustness of DSL to general popularity biases. To this end, we conduct a popularity-filtered evaluation with the head and tail items, which account for the top and bottom 20\% of items based on training interaction counts. In each case, we restrict our test set to include only the positives that fall within these buckets, and average scores over all users with at least one test positive in the bucket. Figure \ref{fig-head_tail} shows the results and demonstrates a consistent tendency for DSL to deliver larger improvements than SL on tail and head items. The Tail--Head line in the same figure highlights the shift in where the gains accrue. It is apparent that the majority of the performance gain for DSL comes from better recommendation of rare, less popular items, with the accuracy improvement on the dominant items being significantly smaller.
This is a very useful property, as historically these items are harder to model and more sensitive to exposure effects. By reshaping the gradient signal, DSL can emphasise informative, competitive alternatives and reduce the influence of abundant (popular) negatives that tend to reinforce popularity.

\section{Conclusion and Future Work}
In this work, we introduced Dual-scale Softmax Loss (DSL), rethinking SL from a competition-first perspective. By reshaping how negatives compete within each training instance, and adapting sharpness between instances, DSL delivered gains consistently over SL and other baseline losses for several backbones. These gains were more pronounced under OOD popularity shift, suggesting improved robustness to unknown distributional biases. Overall, DSL preserves the simplicity and scalability of SL, but makes its gradient budget more metric-relevant, aligning better with the competitors that actually matter for Top-K recommendation. Future work should extend DSL to incorporate uncertainty-aware weighting and alternative sampling strategies to handle exposure bias.
\vspace{-4pt}
\appendix
\section{Optimal Hyperparameters and Additional Results}
\label{appendix}
\begin{table*}[!htbp]
\centering
\begin{tabular}{@{}c@{\hspace{14pt}}c@{}}

\begin{tabular}[t]{llccc}
\toprule
Model & Loss & lr & wd & others \\
\midrule
\multirow{9}{*}{MF}
& BPR        & 0.001 & 0.0001   & -- \\
& GuidedRec  & 0.01  & 0        & -- \\
& LLPAUC     & 0.1   & 0        & $\{0.7, 0.01\}$ \\
& SL         & 0.1   & 0        & $\{0.25\}$ \\
& AdvInfoNCE & 0.1   & 0        & $\{0.2\}$ \\
& BSL        & 0.1   & 0        & $\{0.2, 0.2\}$ \\
& PSL        & 0.1   & 0        & $\{0.1\}$ \\
& SL@K       & 0.1   & 0        & $\{0.25, 2.5, 5\}$ \\
\midrule
\multirow{9}{*}{LightGCN}
& BPR        & 0.001 & 0.000001 & -- \\
& GuidedRec  & 0.01  & 0        & -- \\
& LLPAUC     & 0.1   & 0        & $\{0.7, 0.1\}$ \\
& SL         & 0.1   & 0        & $\{0.2\}$ \\
& AdvInfoNCE & 0.1   & 0        & $\{0.2\}$ \\
& BSL        & 0.1   & 0        & $\{0.05, 0.2\}$ \\
& PSL        & 0.1   & 0        & $\{0.1\}$ \\
& SL@K       & 0.01  & 0        & $\{0.2, 2.5, 20\}$ \\
\bottomrule
\end{tabular}

&

\begin{tabular}[t]{llccc}
\toprule
Model & Loss & lr & wd & others \\
\midrule
\multirow{9}{*}{MF}
& BPR        & 0.001 & 0.00001  & -- \\
& GuidedRec  & 0.01  & 0        & -- \\
& LLPAUC     & 0.1   & 0        & $\{0.5, 0.01\}$ \\
& SL         & 0.01  & 0        & $\{0.2\}$ \\
& AdvInfoNCE & 0.1   & 0        & $\{0.2\}$ \\
& BSL        & 0.1   & 0        & $\{0.5, 0.2\}$ \\
& PSL        & 0.01  & 0        & $\{0.1\}$ \\
& SL@K       & 0.01  & 0        & $\{0.2, 2.25, 20\}$ \\
\midrule
\multirow{9}{*}{LightGCN}
& BPR        & 0.01  & 0.000001 & -- \\
& GuidedRec  & 0.01  & 0        & -- \\
& LLPAUC     & 0.1   & 0        & $\{0.5, 0.01\}$ \\
& SL         & 0.01  & 0        & $\{0.2\}$ \\
& AdvInfoNCE & 0.01  & 0        & $\{0.2\}$ \\
& BSL        & 0.01  & 0        & $\{0.2, 0.2\}$ \\
& PSL        & 0.01  & 0        & $\{0.1\}$ \\
& SL@K       & 0.01  & 0        & $\{0.2, 2.25, 5\}$ \\
\bottomrule
\end{tabular}

\end{tabular}%
\caption{IID Hyperparameters for Amazon-Health (left) and Amazon-Electronic (right). SL@K denotes SL@20.}
\vspace{-25pt}
\label{tab:hyp_health_elec}
\end{table*}

\begin{table*}[!htbp]
\centering
\begin{tabular}{@{}c@{\hspace{14pt}}c@{}}

\begin{tabular}[t]{llccc}
\toprule
Model & Loss & lr & wd & others \\
\midrule
\multirow{9}{*}{MF}
& BPR        & 0.001 & 0.000001 & -- \\
& GuidedRec  & 0.001 & 0        & -- \\
& LLPAUC     & 0.1   & 0        & $\{0.7, 0.01\}$ \\
& SL         & 0.1   & 0        & $\{0.1\}$ \\
& AdvInfoNCE & 0.1   & 0        & $\{0.1\}$ \\
& BSL        & 0.1   & 0        & $\{0.2, 0.1\}$ \\
& PSL        & 0.1   & 0        & $\{0.05\}$ \\
& SL@K       & 0.01  & 0        & $\{0.1, 1, 20\}$ \\
\midrule
\multirow{9}{*}{LightGCN}
& BPR        & 0.001 & 0        & -- \\
& GuidedRec  & 0.001 & 0        & -- \\
& LLPAUC     & 0.1   & 0        & $\{0.7, 0.01\}$ \\
& SL         & 0.1   & 0        & $\{0.1\}$ \\
& AdvInfoNCE & 0.1   & 0        & $\{0.1\}$ \\
& BSL        & 0.1   & 0        & $\{0.05, 0.1\}$ \\
& PSL        & 0.1   & 0        & $\{0.05\}$ \\
& SL@K       & 0.01  & 0        & $\{0.1, 0.75, 5\}$ \\
\bottomrule
\end{tabular}

&

\begin{tabular}[t]{llccc}
\toprule
Model & Loss & lr & wd & others \\
\midrule
\multirow{6}{*}{MF}
& BPR        & $10^{-3}$ & $10^{-6}$ & -- \\
& GuidedRec  & 0.001 & 0        & -- \\
& LLPAUC     & $10^{-1}$ & 0         & $\{0.7, 0.01\}$ \\
& AdvInfoNCE & $10^{-1}$ & 0         & $\{0.05\}$ \\
& SL         & $10^{-1}$ & 0         & $\{0.05\}$ \\
& BSL        & $10^{-2}$ & 0         & $\{0.25, 0.05\}$ \\
& PSL        & $10^{-1}$ & 0         & $\{0.05\}$ \\
& SL@K       & $10^{-1}$ & 0         & $\{0.05, 0.75, 20\}$\\
\midrule
\multirow{6}{*}{LightGCN}
& BPR        & $10^{-3}$ & 0         & -- \\
& GuidedRec  & 0.001 & 0        & -- \\
& LLPAUC     & $10^{-1}$ & 0         & $\{0.7, 0.01\}$ \\
& AdvInfoNCE & $10^{-1}$ & 0         & $\{0.05\}$ \\
& SL         & $10^{-1}$ & 0         & $\{0.05\}$ \\
& BSL        & $10^{-1}$ & 0         & $\{0.025, 0.05\}$ \\
& PSL        & $10^{-1}$ & 0         & $\{0.05\}$ \\
& SL@K       & $10^{-1}$ & 0         & $\{0.05, 0.75, 20\}$ \\
\bottomrule
\end{tabular}

\end{tabular}%
\caption{IID Hyperparameters for Gowalla (left) and Amazon-Movie (right). SL@K denotes SL@20.}
\vspace{-20pt}
\label{tab:hyp_gowalla_movie}
\end{table*}

\begin{figure}[htbp]
\centering
\includegraphics[width=\columnwidth]{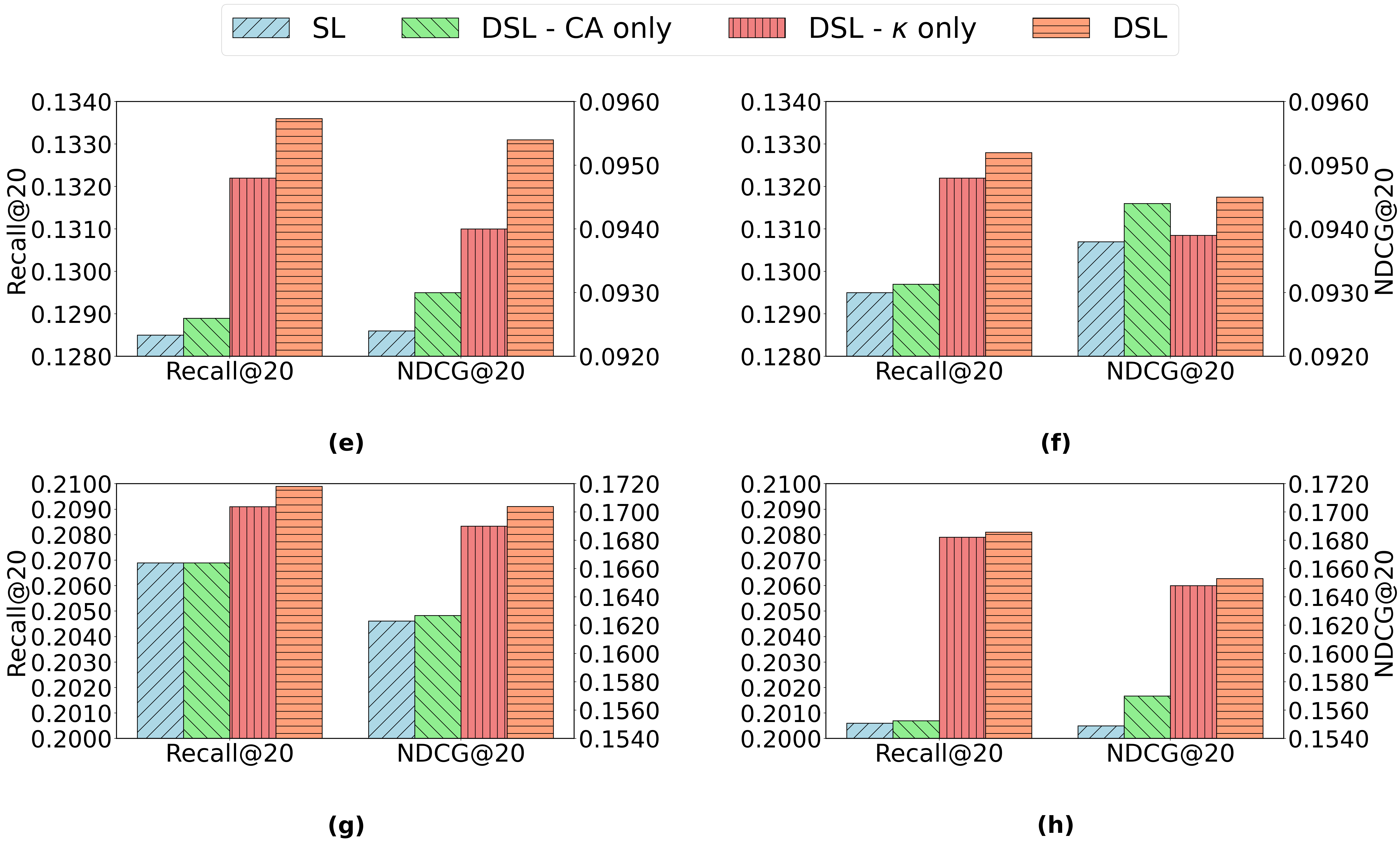}
\vspace{-10pt}
\caption {Movie and Gowalla Ablation}
\label{fig-movie_gowalla_ablation}
\end{figure}

\begin{figure}[htbp]
\centering
\includegraphics[width=0.9\columnwidth]{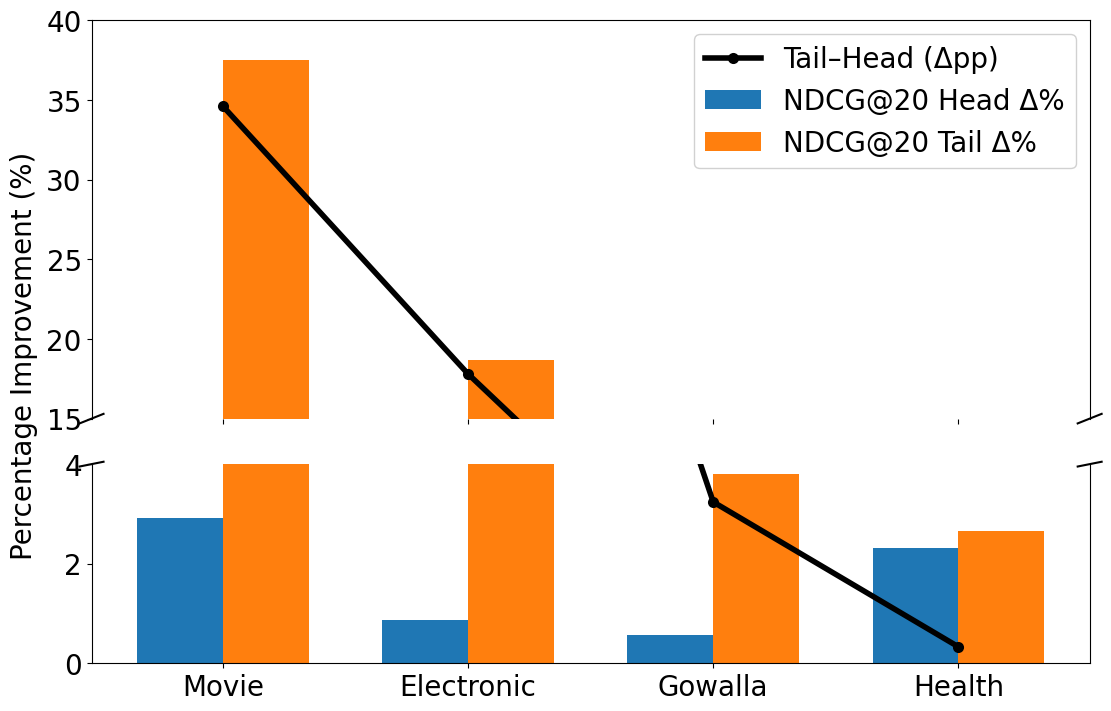}
\vspace{-10pt}
\caption {Percentage improvements for DSL over SL on NDCG@20 for Head and Tail item buckets on LightGCN}
\label{fig-head_tail_light}
\end{figure}

\clearpage
\bibliographystyle{ACM-Reference-Format}
\bibliography{sample-base}

@String{Computer = "{IEEE} Computer" }

@inproceedings{Hu2008ICDM_ImplicitMF,
  title={Collaborative filtering for implicit feedback datasets},
  author={Hu, Yifan and Koren, Yehuda and Volinsky, Chris},
  booktitle={2008 Eighth IEEE international conference on data mining},
  pages={263--272},
  year={2008},
  organization={Ieee}
}

@inproceedings{He2017WWW_NCF,
  title={Neural collaborative filtering},
  author={He, Xiangnan and Liao, Lizi and Zhang, Hanwang and Nie, Liqiang and Hu, Xia and Chua, Tat-Seng},
  booktitle={Proceedings of the 26th international conference on world wide web},
  pages={173--182},
  year={2017}
}

@article{rendle2012bpr,
  title={BPR: Bayesian personalized ranking from implicit feedback},
  author={Rendle, Steffen and Freudenthaler, Christoph and Gantner, Zeno and Schmidt-Thieme, Lars},
  journal={arXiv preprint arXiv:1205.2618},
  year={2012}
}

@article{wu2024effectiveness,
  title={On the effectiveness of sampled softmax loss for item recommendation},
  author={Wu, Jiancan and Wang, Xiang and Gao, Xingyu and Chen, Jiawei and Fu, Hongcheng and Qiu, Tianyu},
  journal={ACM Transactions on Information Systems},
  volume={42},
  number={4},
  pages={1--26},
  year={2024},
  publisher={ACM New York, NY}
}

@inproceedings{wu2024bsl,
  title={Bsl: Understanding and improving softmax loss for recommendation},
  author={Wu, Junkang and Chen, Jiawei and Wu, Jiancan and Shi, Wentao and Zhang, Jizhi and Wang, Xiang},
  booktitle={2024 IEEE 40th International Conference on Data Engineering (ICDE)},
  pages={816--830},
  year={2024},
  organization={IEEE}
}

@article{yang2024psl,
  title={PSL: Rethinking and Improving Softmax Loss from Pairwise Perspective for Recommendation},
  author={Yang, Weiqin and Chen, Jiawei and Xin, Xin and Zhou, Sheng and Hu, Binbin and Feng, Yan and Chen, Chun and Wang, Can},
  journal={Advances in Neural Information Processing Systems},
  volume={37},
  pages={120974--121006},
  year={2024}
}

@inproceedings{yang2025breaking,
  title={Breaking the top-k barrier: Advancing top-k ranking metrics optimization in recommender systems},
  author={Yang, Weiqin and Chen, Jiawei and Zhang, Shengjia and Wu, Peng and Sun, Yuegang and Feng, Yan and Chen, Chun and Wang, Can},
  booktitle={Proceedings of the 31st ACM SIGKDD Conference on Knowledge Discovery and Data Mining V. 2},
  pages={3542--3552},
  year={2025}
}

@article{oord2018CPCrepresentation,
  title={Representation learning with contrastive predictive coding},
  author={Oord, Aaron van den and Li, Yazhe and Vinyals, Oriol},
  journal={arXiv preprint arXiv:1807.03748},
  year={2018}
}

@article{jarvelin2002cumulated_NDCG,
  title={Cumulated gain-based evaluation of IR techniques},
  author={J{\"a}rvelin, Kalervo and Kek{\"a}l{\"a}inen, Jaana},
  journal={ACM Transactions on Information Systems (TOIS)},
  volume={20},
  number={4},
  pages={422--446},
  year={2002},
  publisher={ACM New York, NY, USA}
}

@inproceedings{marlin2009collaborative_MNAR,
  title={Collaborative prediction and ranking with non-random missing data},
  author={Marlin, Benjamin M and Zemel, Richard S},
  booktitle={Proceedings of the third ACM conference on Recommender systems},
  pages={5--12},
  year={2009}
}

@article{mcfadden1972conditional,
  title={Conditional logit analysis of qualitative choice behavior},
  author={McFadden, Daniel},
  year={1972}
}

@article{koren2009matrix,
  title={Matrix factorization techniques for recommender systems},
  author={Koren, Yehuda and Bell, Robert and Volinsky, Chris},
  journal={Computer},
  volume={42},
  number={8},
  pages={30--37},
  year={2009},
  publisher={IEEE}
}

@inproceedings{he2020lightgcn,
  title={Lightgcn: Simplifying and powering graph convolution network for recommendation},
  author={He, Xiangnan and Deng, Kuan and Wang, Xiang and Li, Yan and Zhang, Yongdong and Wang, Meng},
  booktitle={Proceedings of the 43rd International ACM SIGIR conference on research and development in Information Retrieval},
  pages={639--648},
  year={2020}
}

@inproceedings{kang2018self_sasrec,
  title={Self-attentive sequential recommendation},
  author={Kang, Wang-Cheng and McAuley, Julian},
  booktitle={2018 IEEE international conference on data mining (ICDM)},
  pages={197--206},
  year={2018},
  organization={IEEE}
}

@article{saito2019towards_mnar,
  title={Towards resolving propensity contradiction in offline recommender learning},
  author={Saito, Yuta and Nomura, Masahiro},
  journal={arXiv preprint arXiv:1910.07295},
  year={2019}
}

@inproceedings{schnabel2016recommendations_treatments,
  title={Recommendations as treatments: Debiasing learning and evaluation},
  author={Schnabel, Tobias and Swaminathan, Adith and Singh, Ashudeep and Chandak, Navin and Joachims, Thorsten},
  booktitle={international conference on machine learning},
  pages={1670--1679},
  year={2016},
  organization={PMLR}
}

@inproceedings{bruch2019analysis,
  title={An analysis of the softmax cross entropy loss for learning-to-rank with binary relevance},
  author={Bruch, Sebastian and Wang, Xuanhui and Bendersky, Michael and Najork, Marc},
  booktitle={Proceedings of the 2019 ACM SIGIR international conference on theory of information retrieval},
  pages={75--78},
  year={2019}
}

@article{donsker1975asymptotic,
  title={Asymptotic evaluation of certain Markov process expectations for large time, II},
  author={Donsker, Monroe D and Varadhan, SR Srinivasa},
  journal={Communications on Pure and Applied Mathematics},
  volume={28},
  number={2},
  pages={279--301},
  year={1975},
  publisher={Wiley Online Library}
}

@article{namkoong2016stochastic,
  title={Stochastic gradient methods for distributionally robust optimization with f-divergences},
  author={Namkoong, Hongseok and Duchi, John C},
  journal={Advances in neural information processing systems},
  volume={29},
  year={2016}
}

@book{BoydVandenberghe2004,
  title={Convex Optimization},
  author={Boyd, Stephen and Vandenberghe, Lieven},
  publisher={Cambridge University Press},
  year={2004}
}

@inproceedings{zhai2021doro,
  title={Doro: Distributional and outlier robust optimization},
  author={Zhai, Runtian and Dan, Chen and Kolter, Zico and Ravikumar, Pradeep},
  booktitle={International Conference on Machine Learning},
  pages={12345--12355},
  year={2021},
  organization={PMLR}
}

@article{liu2023geometry,
  title={Geometry-calibrated dro: Combating over-pessimism with free energy implications},
  author={Liu, Jiashuo and Wu, Jiayun and Wang, Tianyu and Zou, Hao and Li, Bo and Cui, Peng},
  journal={arXiv preprint arXiv:2311.05054},
  year={2023}
}

@article{wainwright2008graphical,
  title={Graphical models, exponential families, and variational inference},
  author={Wainwright, Martin J and Jordan, Michael I and others},
  journal={Foundations and Trends{\textregistered} in Machine Learning},
  volume={1},
  number={1--2},
  pages={1--305},
  year={2008},
  publisher={Now Publishers, Inc.}
}

@article{burges2006learning,
  title={Learning to rank with nonsmooth cost functions},
  author={Burges, Christopher and Ragno, Robert and Le, Quoc},
  journal={Advances in neural information processing systems},
  volume={19},
  year={2006}
}

@inproceedings{shi2024lower,
  title={Lower-left partial auc: An effective and efficient optimization metric for recommendation},
  author={Shi, Wentao and Wang, Chenxu and Feng, Fuli and Zhang, Yang and Wang, Wenjie and Wu, Junkang and He, Xiangnan},
  booktitle={Proceedings of the ACM Web Conference 2024},
  pages={3253--3264},
  year={2024}
}

@article{zhang2023empowering,
  title={Empowering collaborative filtering with principled adversarial contrastive loss},
  author={Zhang, An and Sheng, Leheng and Cai, Zhibo and Wang, Xiang and Chua, Tat-Seng},
  journal={Advances in Neural Information Processing Systems},
  volume={36},
  pages={6242--6266},
  year={2023}
}

\end{document}